\pdfoutput=1
\documentclass[letterpaper, 10 pt, conference]{ieeeconf}

\usepackage{times}
\usepackage{color, colortbl}
\usepackage[dvipsnames]{xcolor}
\usepackage[colorlinks=true,linkcolor=black,anchorcolor=black,citecolor=black,filecolor=black,menucolor=black,runcolor=black,urlcolor=black]{hyperref}
\usepackage{amsfonts}
\usepackage{amsmath}
\usepackage{amssymb}
\usepackage{dsfont}
\usepackage{subcaption}
\usepackage{multicol,multirow}
\usepackage[pdftex]{graphicx}   
\usepackage{graphicx}
\usepackage[acronyms, shortcuts]{glossaries}
\usepackage[capitalize,noabbrev,nameinlink]{cleveref}
\crefformat{equation}{(#2#1#3)}
\Crefname{figure}{Fig.}{Fig.}
\Crefname{algocfline}{Algorithm}{Algorithms}
\Crefname{algocf}{line}{lines}
\Crefname{assumption}{Assumption}{Assumptions}
\crefrangeformat{assumption}{Assumptions~#3#1#4-#5#2#6}
\usepackage[ruled,vlined,linesnumbered]{algorithm2e}
\definecolor{princetonorange}{RGB}{232,120,34}

\SetCommentSty{mycommfont}
\SetKwInput{KwInput}{Input}                
\SetKwInput{KwOutput}{Output}              
\usepackage{cases}
\usepackage{xspace}
\usepackage{booktabs}
\usepackage{siunitx}
\usepackage{enumerate}
\usepackage[backend=biber,style=numeric-comp,sorting=none,maxbibnames=6]{biblatex}
\usepackage{balance}

\newcommand{\algoName}{SSTA\xspace} 
\usepackage{tikz}
\usepackage[export]{adjustbox}
\usetikzlibrary{arrows,decorations.pathmorphing,backgrounds,fit,positioning,shapes.symbols,chains}
\usetikzlibrary{bayesnet}
\usepackage{bm}
\usepackage{graphicx}

\usepackage{algorithm2e}
\usepackage{outlines}
\let\labelindent\relax
\usepackage{enumitem}
\setenumerate[2]{label=\alph*.}
\setenumerate[3]{label=\roman*.}

\IEEEoverridecommandlockouts
\bibliography{references}

\graphicspath{{./figures/}}    
\DeclareGraphicsExtensions{.pdf,.png,.jpg,.jpeg}

\usepackage[textfont=md,font=footnotesize]{caption}

\usepackage{ifthen}
\newboolean{include-notes}
\setboolean{include-notes}{true}

\usepackage{lipsum}

\newcommand{\david}[1]%
    {\textcolor{orange}{\textbf{DFK: #1}}}
    
\newcommand{\shrey}[1]%
    {\textcolor{teal}{\textbf{SK: #1}}}
    
\newcommand{\jiankai}[1]%
    {\textcolor{magenta}{\textbf{JS: #1}}}
    
\newcommand{\mac}[1]%
    {\textcolor{red}{\textbf{MS: #1}}}

\newcommand{\example}[1]%
{
\textbf{Running example:}
\textit{#1}
}
\glsdisablehyper
\newacronym{ssta}{SSTA}{Self-Supervised Traffic Advisor}
\newacronym{mpc}{MPC}{Model Predictive Control}
\newacronym{mse}{MSE}{Mean Squared Error}
\newacronym{cav}{CAV}{Connected and Autonomous Vehicle}
\newacronym[longplural={Fields of View}]{fov}{FOV}{Field of View}

\newcommand{\mc}{\mathcal}
\newcommand{\mbf}{\mathbf}
\newcommand{\mbb}{\mathbb}
\newcommand{\ts}[1]{\textsuperscript{#1}}
\newcommand{\regtext}[1]{\mathrm{\textnormal{#1}}}
\newcommand{\ith}{$i$\ts{th} }

\newcommand{\kth}{$k$\ts{th} }

\newcommand{\R}{\mbb{R}}
\newcommand{\N}{\mbb{N}}

\newcommand{\norm}[1]{\left\Vert#1\right\Vert}

\newcommand{\nrows}{H}
\newcommand{\ncols}{W}
\newcommand{\nssta}{N}
\newcommand{\ndata}{D}

\newcommand{\tvar}{t}
\newcommand{\horizon}{T}
\newcommand{\hdn}{\mbf{h}}
\newcommand{\im}{\mbf{x}}
\newcommand{\pred}{\hat{\mbf{x}}} 
\newcommand{\msg}{\mbf{y}}
\newcommand{\nbhd}{K}

\newcommand{\loss}{\mc{L}}
\newcommand{\dataset}{\mc{X}}
\newcommand{\predset}{\hat{\mc{X}}}
\newcommand{\msgset}{\mc{Y}}
\newcommand{\predfun}{f}
\newcommand{\param}{\theta}
\newcommand{\paramset}{\R^p}

\newcommand{\alldata}{\mbb{X}}

\newcommand{\allmsg}{\mbb{Y}}

\newcommand{\lbl}[1]{_{\regtext{#1}}}
\newcommand{\frob}{\lbl{F}}

\title{\LARGE \bf Self-Supervised Traffic Advisors: Distributed, Multi-view Traffic Prediction for Smart Cities}
\author{
Jiankai Sun$^1$, Shreyas Kousik$^1$, David Fridovich-Keil$^2$, and Mac Schwager$^1$
\thanks{
$^1$ Dept. of Aeronautics and Astronautics, Stanford University. 
$^2$ Dept. of Aerospace Engineering and Engineering Mechanics, UT Austin.
Corresponding author: \texttt{jksun@stanford.edu}.
Toyota Research Institute provided funds to support this work.}
}

\begin{document}

\maketitle

\begin{abstract}
\acp{cav} are becoming more widely deployed, but it is unclear how to best deploy smart infrastructure to maximize their capabilities.
One key challenge is to ensure \acp{cav} can reliably perceive other agents, especially occluded ones.
A further challenge is the desire for smart infrastructure to be autonomous and readily scalable to wide-area deployments, similar to modern traffic lights.
The present work proposes the \ac{ssta}, an infrastructure edge device concept that leverages self-supervised video prediction in concert with a communication and co-training framework to enable autonomously predicting traffic throughout a smart city.
An \ac{ssta} is a statically-mounted camera that overlooks an intersection or area of complex traffic flow that predicts traffic flow as future video frames and learns to communicate with neighboring \acp{ssta} to enable predicting traffic before it appears in the \ac{fov}.
The proposed framework aims at three goals: (1) inter-device communication to enable high-quality predictions, (2) scalability to an arbitrary number of devices, and (3) lifelong online learning to ensure adaptability to changing circumstances.
Finally, an \ac{ssta} can broadcast its future predicted video frames directly
as information for \acp{cav} to run their own post-processing for the purpose of control.
\end{abstract}

\section{Introduction}\label{sec:intro}

\begin{figure}
    \centering
    \subfloat[Road map]{\includegraphics[width=0.467\linewidth]{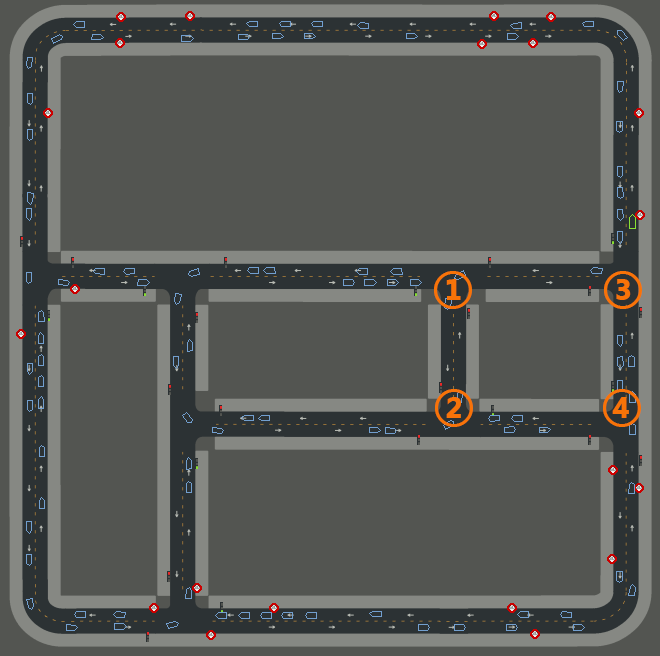}\label{}}
    \hspace{0.01cm}
    \subfloat[Top-down view of the whole scene]{\includegraphics[width=0.49\linewidth]{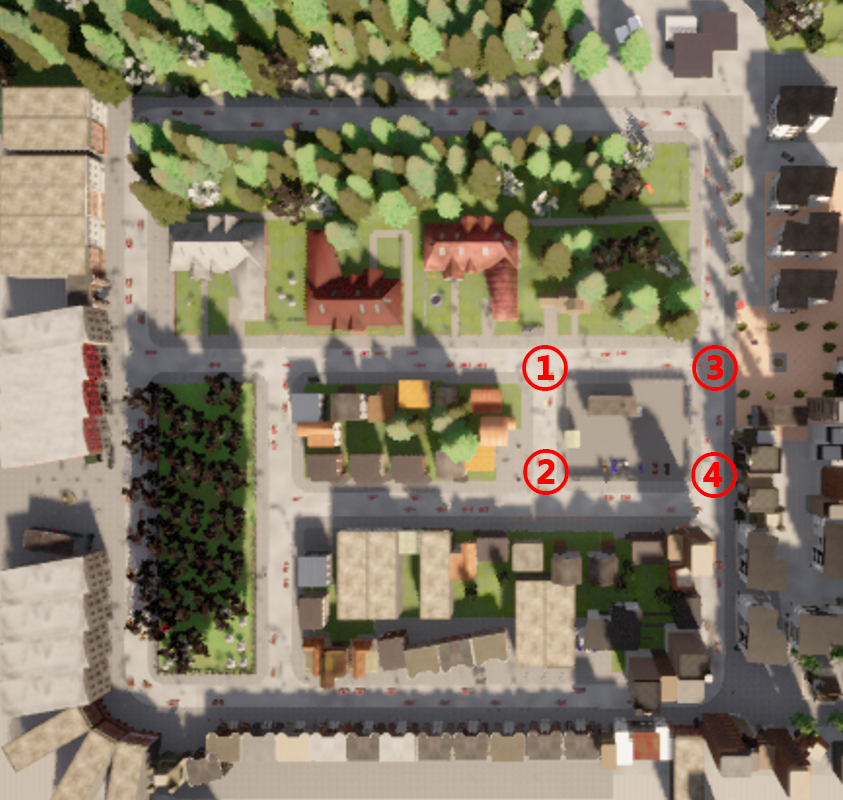}\label{}}
    \vspace{0.1cm}
    \subfloat[View 1]{\includegraphics[width=0.24\linewidth]{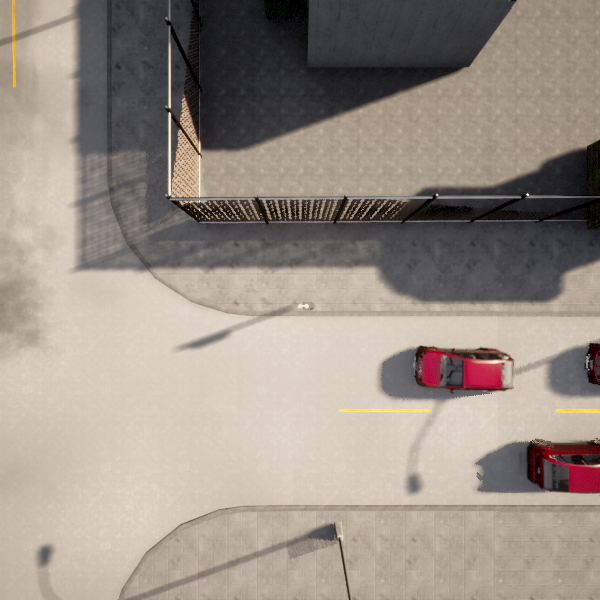}\label{}}
    \hspace{0.001cm}
    \subfloat[View 2]{\includegraphics[width=0.24\linewidth]{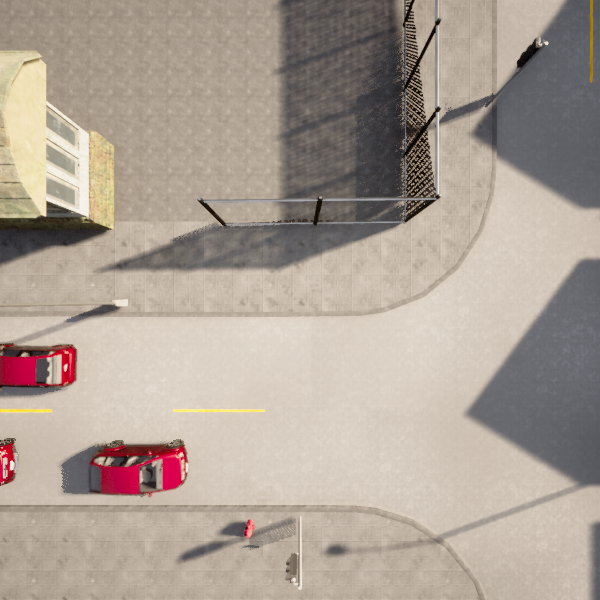}\label{}}
    \hspace{0.001cm}
    \subfloat[View 3]{\includegraphics[width=0.24\linewidth]{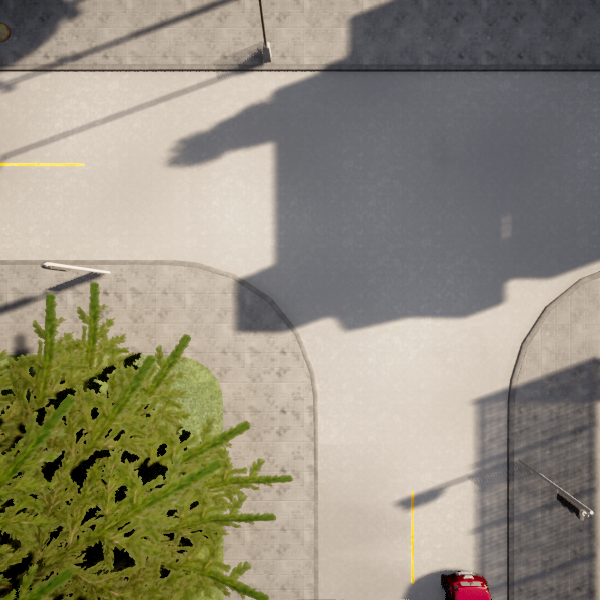}\label{}}
    \hspace{0.001cm}
    \subfloat[View 4]{\includegraphics[width=0.24\linewidth]{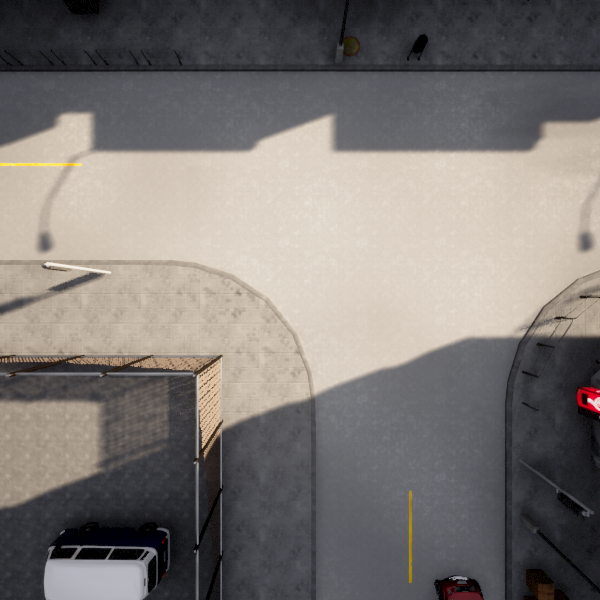}\label{}}
    \caption{\textbf{SSTA Overview:} Multiple
intersections are connected according to the graph induced
by a road network. At each intersection, a camera monitors
the flow of traffic across the intersection, and the data is
processed by an SSTA. Locations of each view are annotated in (a) and (b).}
    \label{fig:teaser}
\end{figure}

Connected and Autonomous Vehicles (CAVs) in smart cities have the potential to drastically improve road safety and traffic throughput.
However, this potential is limited by the challenges of reliably perceiving other agents and predicting how they may move in complicated scenarios, such as unprotected left turns or pedestrian-dense areas.
These challenges are compounded by varying lighting and weather, and by a lack of universal protocols for data sharing between CAVs and smart cities.
To begin addressing these challenges, we propose a novel network of learning-enabled edge computers that broadcast universally accessible data about future occupied space at safety-critical locations such as busy intersections and hidden drives.
We call this a Self-Supervised Traffic Advisor (SSTA), which is a statically mounted camera, paired with a small computer, overlooking a safety-critical location (see~\cref{fig:teaser}).

We envision an SSTA network providing advice to CAVs and enabling faster and safer traffic throughput in a fully distributed, decentralized, asynchronous way.
On its own, an SSTA can learn to predict future camera measurements and, consequently, future occupied space within its field-of-view.
By broadcasting time-varying predictions of occupancy to CAVs, SSTAs can advise future actions without requiring two-way communication.
For example, if a CAV is waiting to make a left turn across traffic, but its view of oncoming traffic is occluded, it can passively receive an SSTA's prediction to help decide when to begin its turn.
By networking with each other, SSTAs learn to pass messages that improve each others' predictive abilities.
For example, a message received by an SSTA can inform it of a vehicle that is about to enter its field-of-view (FOV), resulting in a seamless prediction of traffic across devices and physical locations.
Cameras already exist and are extensively deployed in modern transportation systems.
To tackle the challenge of trustworthiness, existing defense and privacy-preserving techniques such as differential privacy (DP)
can be applied in the future to make our approach robust and secure.

\textit{Contribution.}
The present work makes a first step towards realizing an SSTA network by proposing a decentralized, self-supervised traffic prediction architecture.
We study our method's abilities to learn inter-device communication, scale to many devices, and perform lifelong learning.

\section{Related Work}\label{sec:related}

Our proposed method lies at the nexus of several applications of machine learning: video prediction, lifelong learning, and networked learning.
While we note that there is extensive work on prediction, including for traffic occupancy, using supervised learning \cite{itkina2019dynamic,wang2021predrnn,salzmann2020trajectron++}, here we focus on self-supervised learning from unlabeled, raw video.

\emph{Video Prediction.}
In modern video prediction, the objective is to estimate future frames of video given previous frames, typically using deep neural networks  \cite{mathieu2015deep,walker2016_forecast_var_autoenc_stat_img,babaeizadeh2017stochastic}.
This task is attractive from a data collection perspective since it is often amenable to self-supervision.
However, a key challenge is that individual frames are high-dimensional, resulting in a more difficult problem than just processing a single image; this has been tackled by using recurrent architectures \cite{babaeizadeh2017stochastic,xu2018structure} (which we employ in the present work).

Video prediction typically focuses on a single camera, with the goal of either learning visual dynamics \cite{walker2016_forecast_var_autoenc_stat_img} or interpolating video spatially and temporally \cite{lu2017flexible}.
By contrast, the present work is concerned with \emph{many} static, networked cameras, which resembles the surveillance and multi-view use cases in the literature.
In surveillance, video prediction has been paired with background modeling to efficiently detect anomalies \cite{zhang2013background}. 
In the multi-view case, the challenge of a high-dimensional input is augmented, leading to a variety of efforts to interpolate or synthesize new views by leveraging information shared between views \cite{vyas2020multi,pan2020cross}.
In this work, we consider the multi-view case, but from far-apart cameras, and still seek to learn to share information.

\emph{Lifelong Learning.}
Also known as continual, sequential, or incremental learning, this is the process of learning from data that arrive sequentially, with only a small portion of input data available at a time \cite{parisi2019continual}.
The major challenge is that training a model on recent data may significantly impact its performance on past data; i.e., methods seek to ``learn without forgetting'' \cite{delange2021continual}.
Most work in this space is concerned with supervised, task-based settings; while our setting is instead unsupervised, we are still concerned with saving the \emph{best} data to train an \ac{ssta}, since storage space is limited.

There are a variety of approaches to leveraging past data.
For example, one can store a subset of observed past data, and only take gradient steps from new data that do not increase the loss on the past data \cite{lopez2017gradient}.
To avoid the memory and privacy concerns of saving past data, one can alternatively save encoded, low-dimensional representations \cite{li2017learning,Rannen_2017_ICCV}, or restrict gradient steps on new data to be perpendicular to the steps taken on past data \cite{farajtabar2020orthogonal}.
Instead of constraining gradient steps, one can also focus on a model's parameters; for example, one can keep learned parameters close to those trained for previous tasks \cite{lee2017overcoming}, decide which parameters to update via an importance metric \cite{zenke2017continual}. 
In this work, we test a variety of approaches to understand which is best for our problem setting.

\emph{Networked Learning.}
In the present work, we consider a framework that we call \textit{networked co-learning}, where many agents train their own models simultaneously by passing messages and gradient information between one another.
One can view this as a single, large, physically-distributed network; we leverage the local nature of each agent's inputs and outputs to enable training the entire network in a distributed manner with only local information available for each agent.
We now distinguish this setting from two similar paradigms: distributed learning and graph neural networks.

In distributed and federated learning, one seeks to learn a \textit{single} model that is shared among many agents \cite{he2020fedml}.
In the distributed case, one seeks to use multiple distinct workers to train on centralized data; in the federated case, one uses decentralized data, with an emphasis on privacy by avoiding passing raw data on a network.
There are a variety of architectures that seek a \emph{consensus} of a centralized model in this setting \cite{yu2022dinno,bistritz2020distributed}.

Graph neural networks (GNNs) model a function on data that are best represented by a graph \cite{zhou2020graph}.
This can be seen as a generalization of grid-like data structures (e.g., images), leading to insights such as convolutions or attention mechanisms over graph-structured data~\cite{9491972}.
The key difference in the present work is that we consider a network of agents each training its own model (represented as a neural network); we do not seek to learn a model of the network (as a graph), but rather a network of models.

\section{Proposed Method}
\label{sec:method}
We are interested in multi-view traffic prediction. 
Here, the key technical challenge is for each \ac{ssta} to determine which information is useful to communicate with its neighbors in order to improve their collective prediction performance.
This communication is encoded in a learned latent space.
We first define the \emph{Networked Traffic Prediction} problem setup and how to address it using a latent space message-passing approach.
We then discuss how we learn the recursive prediction task and leverage the conjugate relationships between each view.
Finally, we present our algorithms for lifelong learning and continual deployment in the multi-view setting.

\SetKwComment{Comment}{/* }{ */}
\SetKwFor{For}{for}{do}{end\ for for}%
\SetAlgoNoLine
\begin{algorithm}[t]
\DontPrintSemicolon
\caption{\algoName}
\label{alg:ssta}
  \KwIn{Model parameter $\theta^i$ for node $i$, time step $t$, sequence length $T$, total number of nodes $N$}
  \KwOut{Updated model parameter $\theta$ for all nodes}
\While{not converged}{
\For{$t=1,2,\dots, T$}{
\For{$i=1,2,\dots, N$}{
    node $i$ performs local prediction as \cref{eq:recursive_prediction} and broadcast $y^{i}_t$ to other nodes\;
}
}
Update each node by minimizing  \cref{eq:network_loss}\;
}
\Return $\theta$
\end{algorithm}

\begin{figure*}
    \centering
    \includegraphics[width=0.7\linewidth]{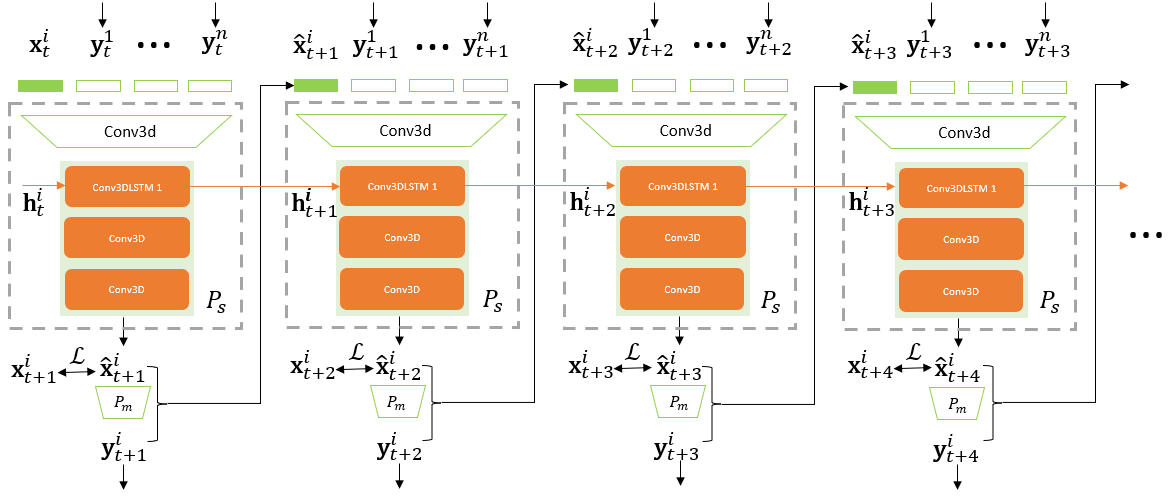}
    \caption{
    The main architecture of a \emph{single} SSTA $i$, in which the orange arrows denote the state transition paths over time. Note that messages $\{\msg_s^j\}, j\ne i, s \ge t$ come from \emph{other} \acp{ssta}. $\mathcal{L}$ is the loss between the prediction $\hat{\im}$ and ground-truth $\im$.}
    \label{fig:single_agent_net}
\end{figure*}

\subsection{Networked Traffic Prediction}

We consider a multi-view image prediction problem that arises in city traffic.
As illustrated in~\cref{fig:teaser}, $\nssta$ intersections are connected according to the graph induced by a road network.
At each intersection, a camera monitors the flow of traffic across the intersection, and the data is processed by an \ac{ssta}.
More precisely, at each time $\tvar$ the \ith \ac{ssta} aims to match a sequence of estimated future images $\predset_\tvar^i$ to the true images $\dataset_\tvar^i$ which are as yet unseen, where
\begin{align}\label{eq:dataset}
    \dataset_\tvar^i = \left\{\im_\tau^i \right\}_{\tau = \tvar}^{\tvar+\horizon}\ \regtext{and}\ 
    \predset_\tvar^i = \left\{\pred_\tau^i \right\}_{\tau = \tvar+1}^{\tvar+\horizon},
\end{align}
with $\im_\tau^i \in \{0,1, \dots,255\}^{\nrows\times\ncols\times 3}$ denoting an RGB image of size $\nrows\times\ncols$, and $\pred^i_\tau$ a prediction of the same size.
We denote the set of true image sequences over the entire network by $\alldata_\tvar = \{\dataset_\tvar^i\}_{i=1}^{\nssta}$.
That is, $\alldata_\tvar$ contains the image sequences for all \acp{ssta} for all times $\tau \in \{\tvar,\dots,\tvar+\horizon\}$.

We seek to optimize the image prediction quality at each \ac{ssta}.
While many metrics for image quality exist \cite{mathieu2015deep}, in this work we optimize the \ac{mse} for simplicity.

\subsection{Message Passing}

As discussed in \cref{sec:related}, video predictions are commonly expressed as functions of history produced by a parameterized model (for instance, ${\pred_{\tvar+1} = \predfun(\pred_\tvar; \param)}$ with parameters $\param \in \paramset$).
In this work, we introduce a novel functional architecture that enables each \ac{ssta}'s prediction to depend upon past observations made by others.
In particular, we design the \ith \ac{ssta} to output an encoded \emph{message} $\msg_\tvar^i$ as a real-valued vector of a user-specified dimension that is fixed \emph{a priori}.
Importantly, each \ac{ssta} sends a \emph{single} message $\msg_\tvar^i$ to some of its neighbors. 

If the \ith \ac{ssta} is connected to a set of neighbors indexed by $\nbhd^i \subset \{1,\dots,\nssta\}$, then, at time $\tvar$, it receives a \emph{message set}
\begin{align}
    \msgset_\tvar^i = \{\msg_\tvar^k\}_{k \in \nbhd^i}.
\end{align}
We denote the set of all messages received by all \acp{ssta} at time $\tvar$ as $\allmsg_\tvar= \{\msgset_\tvar^i\}_{i=1}^\nssta$.

\subsection{Recursive Prediction}

The \ith \ac{ssta} must create a sequence of $\horizon$ predictions $\pred_\tvar^i$ as images, which are a high-dimensional output.
To reduce the output dimensionality of our model, we generate predictions recursively, one time step at a time.
To enable this, the \ith \ac{ssta} maintains a hidden state $\hdn_\tvar^i$ (note, architecture implementation details are below in Section \ref{subsec:network_architecture}).
We initialize the hidden state as all zeros at time $0$.

Then, each \ac{ssta} recursively generates predictions and messages while updating its hidden state:
\begin{align}\label{eq:recursive_prediction}
    (\pred_{\tau+1}^i,\hdn_{\tau+1}^i,\msg_{\tau+1}^i) = \begin{cases}
        \predfun(\im_\tvar^i,\hdn_\tvar^i,\msgset_\tvar^i; \param^i) & \tau = \tvar \\
        \predfun(\pred_{\tau}^i,\hdn_{\tau}^i,\msgset_\tau^i; \param^i) & \tau > \tvar, \\
    \end{cases}
\end{align}
where $\param^i$ denotes the trainable parameters of the \ith \ac{ssta}.
Observe that this recursive model rollout depends on only the messages $\msgset_\tvar^i \in \allmsg_\tvar$ available at time $\tvar$; during prediction ($\tau > t$) the \ac{ssta} discards all messages for times greater than $\tvar$.

Also note that, when deployed (i.e., at test time), each \ac{ssta} predicts over a \emph{receding time horizon}.
That is, at time $\tvar$ it generates predictions $\pred^i_\tau$ according to \cref{eq:recursive_prediction} for $\tau \in \{\tvar, \dots, \tvar + \horizon\}$.
At time $\tvar + 1$, however, the process repeats; to initialize $\cref{eq:recursive_prediction}$ in this instance we use the hidden state $\hdn^i_{\tvar + 1}$ which was generated from the previous prediction rollout and discard that rollout's other hidden states from $\tau > 1$.

\subsection{Networked Co-Learning}

As discussed above, we seek to minimize the \ac{mse} between predicted and ground truth image sequences at all \acp{ssta} simultaneously.
The training loss associated to each time $\tvar$ in the training data is
\begin{align}\label{eq:network_loss}
    \loss_\tvar\big(
        \alldata_\tvar,\allmsg_\tvar
    \big) =
        \sum_{i=1}^{\nssta}\bigg(
            \sum_{\tau = \tvar}^{\tvar+\horizon} \norm{\im_\tau^i - \pred_\tau^i}\frob^2
        \bigg),
\end{align}
where ``F'' denotes the Frobenius norm.
Unlike common federated or distributed learning frameworks \cite{li2020federated,kairouz2019review_fed_learn,yu2022dinno}, our goal is not for all of the \acp{ssta} to learn a single, shared model, but rather for each to learn its own model, which we call \emph{networked co-learning}.

To train the \ith \ac{ssta}, with gradient descent, we must compute the gradient of the overall loss with respect to the \ith set of parameters: $\nabla_{\theta^i} \loss_t$.
To do so efficiently, we observe that the \ith \ac{ssta} is only connected to the \acp{ssta} in $\nbhd^i$.
This sparse graphical dependence means that the aforementioned gradients may be computed in tandem with networks' message information, and 
each \ac{ssta} can therefore perform gradient descent independently on an individual loss $\loss_\tvar^i$ given in the inner sum of \cref{eq:network_loss}. 

To do so, the \kth \ac{ssta}, for $k \in \nbhd^i$, can compute its own \textit{message loss} gradient $\nabla_{\msg_\tvar^i} \loss_\tvar^k$ and send it to the \ith \ac{ssta} in response to receiving the message $\msg_\tvar^i$.
Given these partial derivatives, the \ith \ac{ssta} can compute
\begin{align}
    \nabla_{\theta^i} \loss_{\tvar} = \nabla_{\theta^i} \loss^i_{\tvar} + \sum_{k \in \nbhd^i} (\nabla_{\theta^i} \msg_\tvar^i)^\top \nabla_{\msg_\tvar^i} \loss_\tvar^k,
    \label{eq:ssta_grad}
\end{align}
via the chain rule.
The complete networked co-training algorithm for our framework is shown in Alg.~\ref{alg:ssta}.

\subsection{Lifelong Generative Learning and Continual Deployment}

The advantage of using static cameras is that they can constantly train on data collected online. Unfortunately, it is not tractable to store all past data to reuse for training.
A unique aspect of the \ac{ssta} framework is that we can train and deploy the system continually.
We assume an infinite, continual stream of data. At each time step, the system receives a few consecutive samples $\{\im_t^i\}$ of recent traffic images. 
The goal is to continually learn and update the model that minimizes the prediction errors on previously seen and future samples. In other words, it aims at continuously updating and accumulating knowledge. Given an input model with parameters $\theta$, the system at each time step reduces the empirical risk based on the recently received samples. The learning objective of the online system is Eq.~\eqref{eq:network_loss}.

    

\subsection{Network Architecture}
\label{subsec:network_architecture}

The \ac{ssta} associated with each camera view has an image prediction model $P_s$ and a corresponding pre-trained message generator $P_m$, as~\cref{fig:single_agent_net} shows. 

\subsubsection{Image Prediction Model $P_{s}$}
We use the Conv3DLSTM~\cite{shi2015convolutional} architecture to model spatial and temporal information simultaneously. 
We build a network model by stacking several \texttt{Conv3DLSTM} layers, \texttt{LayerNorm} layer, and \texttt{Conv3D} layers to form an encoding-forecasting structure to predict next step state.

\subsubsection{Message Encoder Model $P_m$} We build an auto-encoder (AE) with \texttt{Conv3D} layers and  \texttt{Conv3DTranspose} layers. This AE model is pre-trained by reconstructing a portion of the images from the training set. We use the encoder of this pre-trained AE as the message generator at each time.

\section{Experiments}
\label{sec:experiments}
Our experiments focus on the two core ideas underlying an \ac{ssta} network: 1) networked co-learning for video prediction, and 2) lifelong learning.

\subsection{Datasets}

We evaluate our method on \emph{CARLA multi-view dataset} quantitatively and qualitatively.
Due to the lack of a suitable multi-view traffic dataset, we collected data from the CARLA simulator~\cite{Dosovitskiy17,pmlr-v155-huang21a,pmlr-v155-sun21a}.
Our \emph{CARLA multi-view dataset} records the trajectories of consecutive multi-view traffic flows in default \texttt{Town02} in the form of top-down-view RGB video frames from cameras placed throughout the environment. The size of each top-down-view frame is $128\times 128\times 3$. The training set contains $11.2$k frames from 8 camera locations while the validation set contains $4.8$k frames from the same camera locations but from different times. 

\begin{table*}[t]
\begin{center}
\renewcommand\tabcolsep{0.5pt}
\begin{small}
\begin{sc}
\begin{tabular}{>{\centering\arraybackslash}m{0.06\linewidth}|>{\centering\arraybackslash}m{0.09\linewidth}>{\centering\arraybackslash}m{0.09\linewidth}|>{\centering\arraybackslash}m{0.09\linewidth}>{\centering\arraybackslash}m{0.09\linewidth}|>{\centering\arraybackslash}m{0.09\linewidth}>{\centering\arraybackslash}m{0.09\linewidth}|>{\centering\arraybackslash}m{0.09\linewidth}>{\centering\arraybackslash}m{0.09\linewidth}|>{\centering\arraybackslash}m{0.09\linewidth}>{\centering\arraybackslash}m{0.09\linewidth}}
\toprule
Step $t$ & \multicolumn{2}{c|}{Step $1$} & \multicolumn{2}{c|}{Step $2$}  & \multicolumn{2}{c|}{Step $3$}  & \multicolumn{2}{c|}{Step $4$}  & \multicolumn{2}{c}{Step $5$}  \\
\midrule
View & View $1$ & View $2$ & View $1$ & View $2$ & View $1$ & View $2$ & View $1$ & View $2$ & View $1$ & View $2$  \\
\midrule
GT & \includegraphics[width=\linewidth]{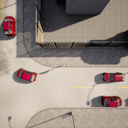} & \includegraphics[width=\linewidth]{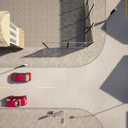}  & \includegraphics[width=\linewidth]{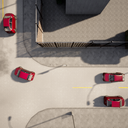} & \includegraphics[width=\linewidth]{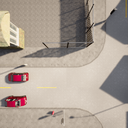}  &
\includegraphics[width=\linewidth]{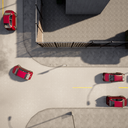} & \includegraphics[width=\linewidth]{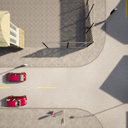}  &
\includegraphics[width=\linewidth]{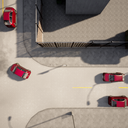} & \includegraphics[width=\linewidth]{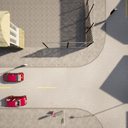}  &
\includegraphics[width=\linewidth]{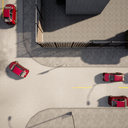} & \includegraphics[width=\linewidth]{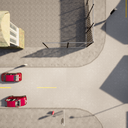}  
\\
\midrule
Pred & \includegraphics[width=\linewidth]{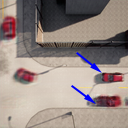} & \includegraphics[width=\linewidth]{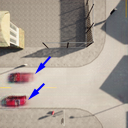}  & \includegraphics[width=\linewidth]{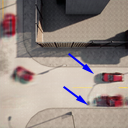} & \includegraphics[width=\linewidth]{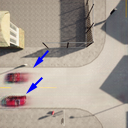}  &
\includegraphics[width=\linewidth]{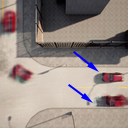} & \includegraphics[width=\linewidth]{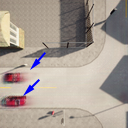}  &
\includegraphics[width=\linewidth]{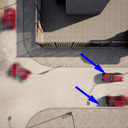} & \includegraphics[width=\linewidth]{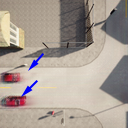}  &
\includegraphics[width=\linewidth]{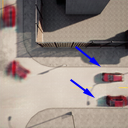} & \includegraphics[width=\linewidth]{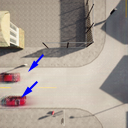}  
\\ 
\bottomrule
\end{tabular}
\end{sc}
\end{small}
\captionof{figure}{\textbf{Qualitative Results}: 2-view prediction. We highlight the cars in the views related to the messages with blue arrows to show the message utility.
}
\label{tab:qual_results_2_view}
\end{center}
\end{table*}

\begin{table*}[t]
\begin{center}
\renewcommand\tabcolsep{0.5pt}
\begin{small}
\begin{sc}
\begin{tabular}{>{\centering\arraybackslash}m{0.06\linewidth}|>{\centering\arraybackslash}m{0.09\linewidth}>{\centering\arraybackslash}m{0.09\linewidth}|>{\centering\arraybackslash}m{0.09\linewidth}>{\centering\arraybackslash}m{0.09\linewidth}|>{\centering\arraybackslash}m{0.09\linewidth}>{\centering\arraybackslash}m{0.09\linewidth}|>{\centering\arraybackslash}m{0.09\linewidth}>{\centering\arraybackslash}m{0.09\linewidth}}
\toprule
Step $t$ & \multicolumn{2}{c|}{Step $1$} & \multicolumn{2}{c|}{Step $2$}  & \multicolumn{2}{c|}{Step $3$}  & \multicolumn{2}{c}{Step $4$}  \\
\midrule
& View $1$ & View $2$ & View $1$ & View $2$ & View $1$ & View $2$ & View $1$ & View $2$ \\
 & \includegraphics[width=\linewidth]{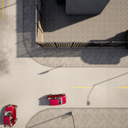} & \includegraphics[width=\linewidth]{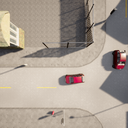}  & \includegraphics[width=\linewidth]{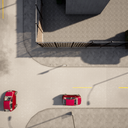} & \includegraphics[width=\linewidth]{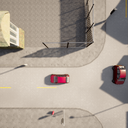}  &
\includegraphics[width=\linewidth]{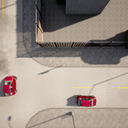} & \includegraphics[width=\linewidth]{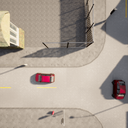}  &
\includegraphics[width=\linewidth]{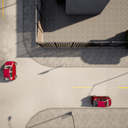} & \includegraphics[width=\linewidth]{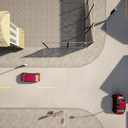}  \\
GT  & View $3$ & View $4$ & View $3$ & View $4$ & View $3$ & View $4$ & View $3$ & View $4$ \\
 & \includegraphics[width=\linewidth]{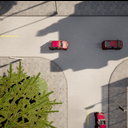} & \includegraphics[width=\linewidth]{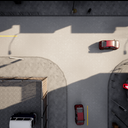}  & \includegraphics[width=\linewidth]{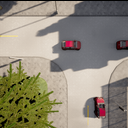} & \includegraphics[width=\linewidth]{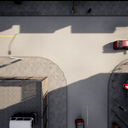}  &
\includegraphics[width=\linewidth]{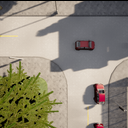} & \includegraphics[width=\linewidth]{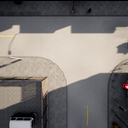}  &
\includegraphics[width=\linewidth]{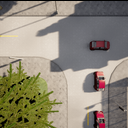} & \includegraphics[width=\linewidth]{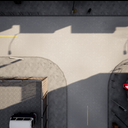}  
\\
\midrule
& View $1$ & View $2$ & View $1$ & View $2$ & View $1$ & View $2$ & View $1$ & View $2$ \\
 & \includegraphics[width=\linewidth]{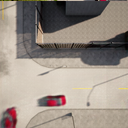} & \includegraphics[width=\linewidth]{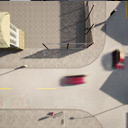}  & \includegraphics[width=\linewidth]{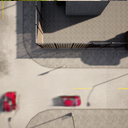} & \includegraphics[width=\linewidth]{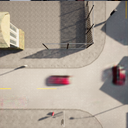}  &
\includegraphics[width=\linewidth]{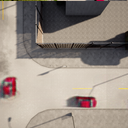} & \includegraphics[width=\linewidth]{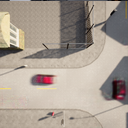}  &
\includegraphics[width=\linewidth]{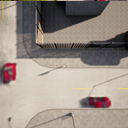} & \includegraphics[width=\linewidth]{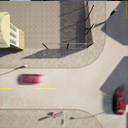}  \\
Pred & View $3$ & View $4$ & View $3$ & View $4$ & View $3$ & View $4$ & View $3$ & View $4$ \\
 & \includegraphics[width=\linewidth]{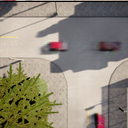} & \includegraphics[width=\linewidth]{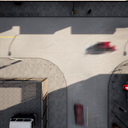}  & \includegraphics[width=\linewidth]{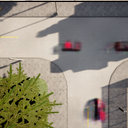} & \includegraphics[width=\linewidth]{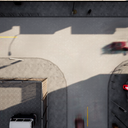}  &
\includegraphics[width=\linewidth]{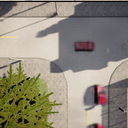} & \includegraphics[width=\linewidth]{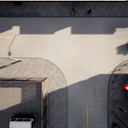}  &
\includegraphics[width=\linewidth]{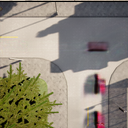} & \includegraphics[width=\linewidth]{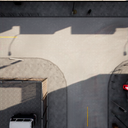}  
\\ 
\bottomrule
\end{tabular}
\end{sc}
\end{small}
\captionof{figure}{\textbf{Qualitative Results}: 4-view prediction.}
\label{tab:qual_results_4_view}
\end{center}
\end{table*}

\subsection{Metrics}
We compare the baselines with the metrics such as Mean Square Error (MSE), Structural Similarity Index Measure (SSIM), Peak Signal to Noise Ratio (PSNR), and Learned Perceptual Image Patch Similarity (LPIPS)~\cite{8578166}.
For PSNR and SSIM, a higher value denotes a better prediction performance. The value of SSIM ranges between -1 and 1, and a larger score means a greater similarity between two images.
The difference between these metrics is that MSE estimates the absolute pixel-wise errors, PSNR/SSIM measures the similarity of structural information within spatial neighborhoods, while LPIPS is based on deep features and aligns more closely to human perceptions.

\subsection{Implementation Details}
We use the ADAM optimizer~\cite{DBLP:journals/corr/KingmaB14} to train the models with a mini-batch of 10 sequences. Unless otherwise specified, we set the learning rate to $10^{-3}$ and stop the training process after 100 epochs. We set the number of channels of each hidden state to 128 and the size of convolutional kernels inside the \texttt{Conv3DLSTM} unit to $5 \times 5$. The quantitative results are averaged over $10$ prediction timesteps.

\subsection{Networked Co-Learning for Video Prediction}
First, we evaluate the multi-view prediction capability of our model. In \cref{tab:qual_results_2_view,tab:qual_results_4_view}, we visualize some examples of the predicted results on the \emph{CARLA multi-view dataset}, where the task is to predict the RGB images and the corresponding emerged messages for each subsequent step. From top to bottom, we see the ground truth, and prediction result for each view at each step. 

Our model is able to predict a sequence of RGB images with 
corresponding emerged messages for multi-agent communication.
Specifically, the most challenging part in multi-view traffic prediction is the bridging of traffic flow ``out'' and ``in'' in different views, where communication with messages plays an important role, and can only be inferred from context and sequence relationships between multiple views.

We further show our model’s performance as the number of views increases. Intuitively, the higher the number of views, the more complex the task, as it involves more messages to communicate. From the results of \cref{tab:quan_results} and \cref{tab:qual_results_2_view,tab:qual_results_4_view}, our framework can effectively predict future traffic not only when the number of views is small (2-view), but also when the number of views is large (4/8-view). This scaling is made possible by \acp{ssta}' ability to learn compact message formats which include only the information relevant for others to improve their predictions.

\begin{table}[]
  \centering
  \begin{tabular}{lcccc}
    \toprule
  {\bf \# Views} & MSE$\downarrow$ & SSIM$\uparrow$ & PSNR$\uparrow$ & LPIPS$\downarrow$ \\
    \midrule
    \texttt{2-view} 
    & $0.16$ & $0.9661$ & $29.64$ & $0.0513$ \\
    \texttt{4-view} 
    & $0.16$ & $0.9625$ & $29.30$ & $0.0535$ \\
    \texttt{8-view} 
    & $0.18$ & $0.9501$ & $28.41$ & $0.0581$\\
    \bottomrule
  \end{tabular}
  \caption{\textbf{Scalability:}
    Performance of our framework \algoName on different number of views.
    }
  \label{tab:quan_results}
\end{table}

\begin{table}[]
  \centering
  \begin{tabular}{lcccc}
    \toprule
  {\bf Method} & MSE$\downarrow$ & SSIM$\uparrow$ & PSNR$\uparrow$ & LPIPS$\downarrow$ \\
    \midrule
    \texttt{SW ($\ndata=50$)} & $0.24$ &$0.8912$ & $23.39$ & $0.1199$ \\
    \texttt{SW ($\ndata=150$)} & $0.22$ &$0.9061$ & $24.59$ & $0.1001$ \\
    \texttt{SW ($\ndata=300$)} & $0.21$ & $0.9175$ & $26.18$ & $0.0801$ \\
    \texttt{\textbf{ID} ($\bm{\ndata=300}$)} & $\bm{0.19}$ & $\bm{0.9481}$ & $\bm{28.35}$ & $\bm{0.0603}$ \\
    \bottomrule
  \end{tabular}
  \caption{\textbf{Lifelong Learning:} influence of different strategies and different sizes of the sliding window $\ndata$. SW: Sliding Window, ID: Interesting Data.
    }
  \label{tab:abl_window_size}
\end{table}


\subsection{Lifelong Learning}
In a practical intelligent transportation application scenario, it will be critical for \acp{ssta} to learn continuously based on streaming input data, and update the model parameters in real-time. 
Hence, we seek to understand the effects of different strategies for lifelong learning on our video prediction task.
We assume an \ac{ssta} can store up to $\ndata \in \N$ time steps worth of data, and compare the following strategies:
\begin{enumerate}
    \item \textit{Sliding Window (SW).}
        At each time $\tvar$, each \ac{ssta} continually trains on the past $\ndata$ time steps of data.
        
    \item \textit{Interesting Data (ID).}
        The ID strategy is modified based on the task-free continual learning framework~\cite{aljundi2019task}. At each time $\tvar$, we save the latest datum for future training if the norm of the gradient $\norm{\nabla_{\param^i} \loss_\tvar^i}_2$ is larger than the average norm of past data gradients. 
\end{enumerate}
We also ablate over different sizes of the sliding window $\ndata$. From the results in \cref{tab:abl_window_size}, we can see that increasing the size of the sliding window $\ndata$ and using the ID training strategy is useful. Quantitative results 
validate the effectiveness of our framework in the 
lifelong learning setting and demonstrate its potential for deployment in traffic scenarios.


\section{Conclusion}
\label{sec:conclusion}

We present \aclp{ssta}, a learning
framework that leverages self-supervised video prediction to autonomously predict traffic throughout a smart city. We argue that communication and co-training are useful for this multi-view prediction. 
Our experimental results show that the \ac{ssta} framework achieves three goals: (1) inter-device communication to enable high-quality predictions, (2) scalability to an arbitrary number of devices, and (3) lifelong online learning to ensure adaptability to changing circumstances.
Our work constitutes a major step towards the goal of enabling multiple autonomous agents to predict real-world streaming data.
Future work will be devoted to improving prediction quality, especially in the transition region in different views, and integrating \acp{ssta} with downstream tasks such as vehicle motion planning in complex traffic scenarios. Our framework
can also be combined with privacy-preserving techniques, such as differential privacy.  
\printbibliography

\end{document}